\documentclass[12pt]{article}

\usepackage[english]{babel}

\usepackage[letterpaper,top=2cm,bottom=2cm,left=3cm,right=3cm,marginparwidth=1.75cm]{geometry}

\usepackage{amsmath}
\usepackage{graphicx}
\usepackage[colorlinks=true, allcolors=blue]{hyperref}

\usepackage{siunitx} 
\title{Stack Index Prediction Using Time-Series Analysis}
\author{Raja CSP Raman (raja@tactii.com) \and Rohith Mahadevan (rohithmahadev30@gmail.com) \and Divya Perumal (divya07msd@gmail.com) \and Vedha Sankar (vedhasankar26@gmail.com) \and Talha Abdur Rahman (cooltalha27@gmail.com)}

\begin{document}
\maketitle

\begin{abstract}
The Prevalence of Community support and engagement for different domains in the tech industry has changed and evolved throughout the years. In this study, we aim to understand, analyze and predict the trends of technology in a scientific manner, having collected data on numerous topics and its growth throughout the years in the past decade. We apply machine learning models on collected data, to understand, analyze and forecast the trends in the advancement of different fields. We show that certain technical concepts such as python, machine learning and Keras have an undisputed uptrend, finally concluding that the Stackindex model forecasts with high accuracy and can be a viable tool for forecasting different tech-domains.
Keywords: Machine Learning, Time Series Forecasting, Tech-Domain Analysis, Kats, Streamlit

\end{abstract}

\section{Introduction}

Software development and maintenance is a complex activity involving many important decisions that need to be made. The choice of programming language and software tool is one such decision. From the perspective of managers and developers of any project, this decision not only affects the performance of the product but also dictates the talent pool and community support available. As far as the developers are concerned, it dictates the current job opportunities and their future career trajectory. We analyse the popularity and the community friendliness of programming languages and technical fields to estimate community engagement and popularity. For our analysis, we look at data from Stack Overflow, one of the most popular programming communities. Stack Overflow is a popular online programming Q and A community providing its participants with rapid access to knowledge and the expertise of their peers. Community support is a valuable tool for developers in any domain. Therefore, a more open, welcoming and responsive (i.e. friendly) community is a blessing which allows us to be more productive as a developer. 
Data such as the questions asked, quality and time-frame of the response is a good indicator of the friendliness of a particular programming community. We take this data on different tech domains and use machine learning models to predict the future numerical. These metrics provide a holistic view of different technologies. We then use these metrics to compare different domains and help answer questions such as: which is the highest trending domain?, which language is most in demand right now?, suggest an alternative language because I work with x language but the community support is bad, etc. The remainder of this paper is organised as follows: we take a look at the literature survey related to our research in Section 2 followed by a Data architecture, understanding of related work in Section 3. In Section 4, we provide a detailed description of the data sets and methodology used. We describe our analytic in Section 5 followed by Result Analysis in Section 6. In Section 7 we do the error analysis and finally we draw our understandings and conclude our paper.

\section{Literature Review}

The major fear of the companies these days is whether the tech stack which is used by their company will survive after 10 years or not. Knowing the growing tech stack is a major factor in developers so that they can upgrade themselves. How do we know which technologies are emerging? Where do we get data of what technologies developers are using the most these days? Stackoverflow is the key.

Stackoverflow is the most active community for developers[1]. On a daily basis around 7000 questions are posted on Stackoverflow by people around the world. Stackoverflow being open source, is used to share knowledge among users. More than 18M questions have been posted in Stackoverflow so far. Each question posted on Stackoverflow will contain tags (keywords) based on the technology on which the question is asked. Using the tags, we can find the number of questions asked by the community for that particular domain.[2]  Based on the tags for each question or query on Stackoverflow we can find the number of questions posted in total. With the help of the number of questions posted can predict the emerging technology. The tag of the technology with the highest count is used most by developers in their day-to-day life. In our project we used this feature of Stackoverflow to analyze and predict the future trends of any domain. [3] The viewers for each question are also important for the growth of technology or languages. For each question posted with a tag, others, apart from the creator of the question, are also considered.

June Young Lee et al, predicted the growth of the technologies with the help of a number of research papers published between 2006 to 2017[4]. Jieun Kim and Christopher L. Magee used the patent citation for eight distinct periods to characterize the structural change and evolutionary behaviors in dynamic technology networks [5].

Won Sang Lee in their paper predicted technologies that will be helpful for predicting medical diseases [6]. They found a convergence pattern among activating catalysts, printing, advanced networking, controlling devices, secured communication with in-memory systems, television systems with pattern recognition, and image processing and analyzing technologies.  ByungunYoon et al in their paper suggested an approach to conducting technology opportunity analysis [7]. They visualized patent information, such as patent documents and citation relationships. This will be very essential in the future for us to identify areas of high end technology. In addition to that Jaewoong Choia et al have examined patents which have high business potential that are used to uncover hidden core technologies of competitive firms and to identify newly issued patents that have a high potential in emerging technologies [8].

As the future is unpredictable, we can say that time series is non-deterministic in nature [10]. Time series is said to be the consistency of time intervals. By defining the interval, it offers the more valuable insights. Time series analysis does a good job of bridging theory and technique. Time series can be used in various fields, one among them is economics[11]. Time series include both discrete time and continuous time models[12]. To forecast the future technologies we used Time Series and Kats algorithms.[13] Time series can represent any number of variables over time. It is also used to increase the efficiency of the model. With efficient models of time series, the future technologies are predicted.The vast majority of time series data is real valued.[14] Another use of time series is Forecasting[15]. To forecast in time series previous year data is must. Hence we have collected data for the past 11 years. Time series analysis is also used to draw insight of time series data points. It computes the changes which will take place over time.

Data mining is the process of extracting and discovering anomalies, patterns and correlations within large data sets to predict the outcome. Time series prediction is just a part of temporal data mining and statistics. It is the process of collecting data at regular intervals to study and predict it. The assumption of time series analysis is to see if the information will repeat it's information in the future. It uses history data to predict the future [16].

Early attempts were made to study time series and it was generally characterized by the idea of a deterministic world. Yule contributed the most in helping the launch of the notion of time series by hypothesizing that that time series can be regarded as the realization of a stochastic process every time. With this idea in mind, people like Slutsky, Walker, Yaglom, and Yule developed the concepts of Auto Regressive model (AR) and Moving Average models(MA) [17].

Prophet is a time series forecasting introduced by Facebook recently [18]. The Prophet uses a decomposable time series model with three main model components: trend, seasonality, and holidays. The Prophet will try to fit the linear and nonlinear functions as components.Predicting  the technology to invest in to prevent the loss of the company similarly here one can improve skill set in emerging technology [20]. It is mostly like predicting the weather of the day and taking precautions. 

\section{Time series Analysis}

Time series analysis is used to obtain data in regular intervals of time to predict what is going to happen in the future. The data is collected in regular intervals of time so that it yields less errors and is sequential in nature. The components or behaviour of time series are trend, seasonality, cycles, and variation. Trend is the long term direction of a series while seasonality is the repeated behaviour in data which occurs at regular intervals. Cycles follow up and down variations with no regularities and variations are completely random and may or may not have irregularities.

For our project we've used the Kats and Prophet models to predict how much engagement a particular technology will have over the next year. We've used data from 2008 to 2019 to forecast the trend of languages. We tried analysing the data using other models like Auto Regressive Integrated Moving Average (ARIMA), Holt Winters and Ensemble, but we will be sticking to Kats and Prophet as it yielded a lesser error rate compared to the others.

Kats is a recent framework released by Facebook and GitHub that is used to analyse and perform time series analysis. Time series analysis can be done when data is collected at regular intervals of time, and is analysed so that we can make a prediction on what the analysis could lead to using machine learning. Kats is useful for Forecasting, Anomaly and Change Point Detection and Feature Extraction. A few forecasting models supported by Kats are Linear, Quadratic, ARIMA, SARIMA, Holt-Winters, Prophet, AR-Net, LSTM, Theta etc. According to Kats' GitHub page, each model follows the sklearn model API pattern. Kats also provides its own set of models to detect outliers, change points and trend changes in the time series data.

Prophet is a forecast of time series data based on an additive model where nonlinear trends fit with yearly, weekly and daily seasonality, in addition to holiday effects. It works best when the time series has strong seasonal effects and several seasons of historical data. Prophet handles missing data and outliers well and adapts well to the shift in trend. Prophet follows the sklearn model API to create it's instances and such.
 
 \section{Data Architecture}

\begin{figure}[!h]
\centering
\includegraphics[width=0.9\textwidth]{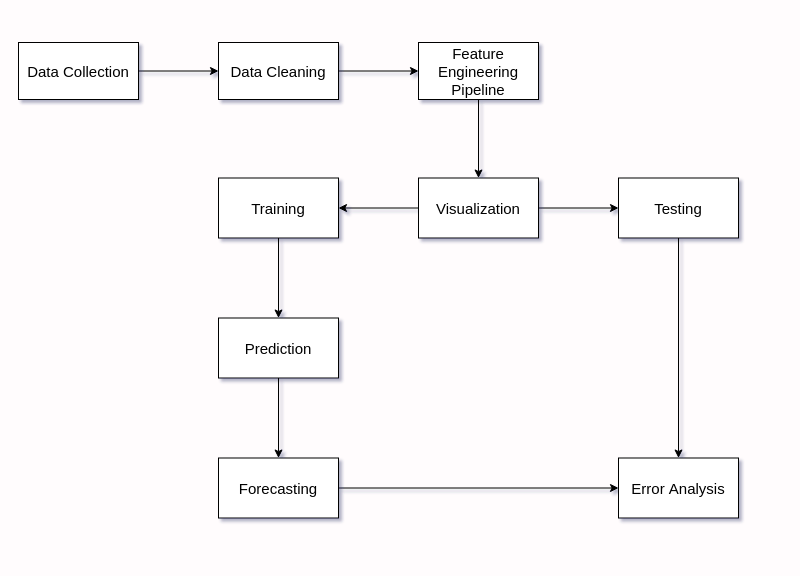}
\caption{\label{fig:data Architecture}Data Architecture - Stack Index Prediction Using Time-Series Analysis}
\end{figure}

\section{Methodology}

In this paper, we are focusing on building time series predictions and analysing the results with the data. The output is the forecasted value. In order to attain the forecasted value, we need a series of steps to be done. This will ensure the data is preprocessed properly and outliers are removed in this case. 

\subsection{Data preparation}

The data is collected from Stackoverflow. Stack exchange queries are used for collecting the data and all the data are stored in comma separated value format. The data that has been collected is processed in python. The data consists of values that are collected from Stackoverflow. As of now, there are 132 rows and 104 columns for the preliminary analysis. In the future we can. The dates are from 2009 to 2019. All the prevailing technologies are listed in the columns. The ultimate aim for our data preparation is to check whether the data is consistent and to convert it into a date object. 

\subsection{Data understanding}

From the preliminary research it is observed that ‘Python’ language has the largest set of questions that is present in stack overflow. The next one is ‘R’ language. In the similar way, we take the top 10 technologies that are frequently used by the community. It is very evident that whenever a particular technology is used doubts and queries may arise. As a result, there will be a lot of questions related to that technology in Stackoverflow. With this method, we subset our data and process the results. In the considerations of community interaction we take the following top 10 technologies. They are Python, R, Pandas, Numpy, Keras, Tensorflow, Machine learning, Deep learning, Apache spark, OpenCV.

\subsection{Data visualization}

Initially, we understood that python is the top technology that is dominating stack overflow. The figure 2 represents that python is fast evolving over time and also lots of queries are posted in python.  This shows that  python has a large community. With this, we can infer that python will be a superior one in future when compared to others. In figure 3, the other technologies which are blooming are visualized. R has a higher interaction next to python. 

\begin{figure}[!htb]
\centering
\includegraphics[width=0.5\textwidth]{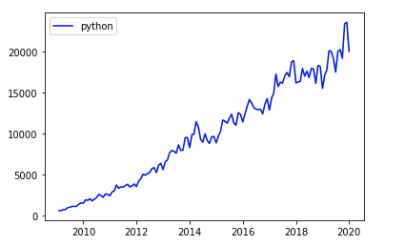}
\caption{\label{fig:python-technology-trend}Python Technology Trend}
\end{figure}

\begin{figure}[!htb]
\centering
\includegraphics[width=0.5\textwidth]{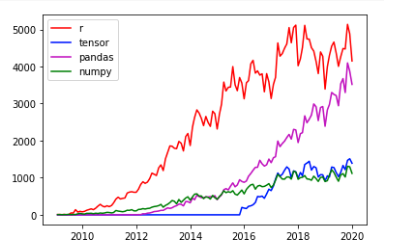}
\caption{\label{fig:other-technologies}Other Technologies}
\end{figure}

Most of these technologies are concerned with machine learning and application development. Hence in order to have a strong foundation we can check whether this thesis is correct. Figure 4 represents that machine learning, deep learning,  OpenCV and Keras are the next blooming technologies. As time passes, the need for the technology also increases. But, this may not be the case with all technologies. Some technologies like flash are going down in trend. 

\begin{figure}[!htb]
\centering
\includegraphics[width=0.5\textwidth]{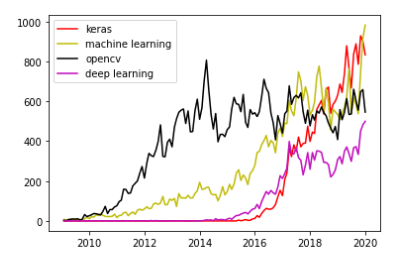}
\caption{\label{fig:trending-technologies}Trending Technologies}
\end{figure}

\begin{figure}[!htb]
\centering
\includegraphics[width=0.5\textwidth]{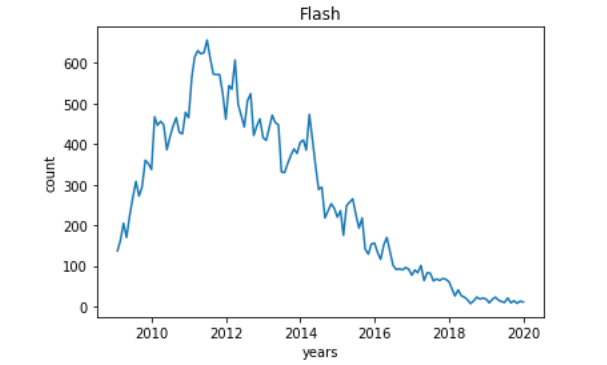}
\caption{\label{fig:flash-technology-downtrend}Flash Technology Downtrend}
\end{figure}

\subsection{Data forecasting}

We will be using time series prediction to forecast the future. Python is used as the primary language for our prediction. Initially, we subset the data set and convert it into a time series data frame. For each column, a separate model can be built. This is because; we can initially see the trend for that particular technology.  The Prophet model from Kats is used for our prediction. Figure 6 represents python forecasting. It illustrates that python will be dominating the other languages in the upcoming months. The reason behind this is the flexibility and compatibility of the technology. 

\begin{figure}[!h]
\centering
\includegraphics[width=0.5\textwidth]{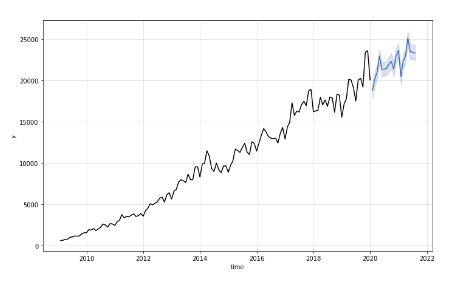}
\caption{\label{fig:python-technology-forecasting}Python Technology Forecasting}
\end{figure}

\section{Analysis of results}

The overall prophet model fits well with the data. We can infer that the trend is different for different languages. We can take one such technology and analyze the trend. Keras is chosen from the top 10 technologies and the forecasting is represented in figure 6. It shows that Keras is going to prevail in the next upcoming years and the need for the technology is high.

\begin{figure}[!h]
\centering
\includegraphics[width=0.5\textwidth]{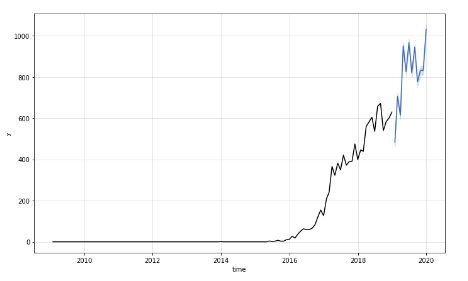}
\caption{\label{fig:keras-forecasting}Keras Forecasting}
\end{figure}

In addition to that top technologies when combined, are also forecasted. We know that, in today’s information world we do not rely on any particular technology. Rather, we use multiple technologies for the betterment of results. Such methodology will lead to increased and efficient use. In the similar way, we have taken some combinations for our analysis. Such a combination includes machine learning with Python, R and Machine learning (theory based). Another one is a set of frameworks. Frameworks that are combined are Keras, Tensorflow and Pytorch. As they are the basic technologies which are used predominantly in the field of deep learning. In addition to that this paper also concentrates on where the threshold point for any particular technology lies on. Figure 7 explains the trend of deep learning framework (Keras, Pytorch and Tensorflow combination) and figure 8 illustrates the increase in change point. This point indicates the deep learning frameworks substantially increased from 2017 . For this detection SARIMA is used as it is considered as a best fit for the data points. The reason for using SARIMA here is because of the data points. The subset contains some data points which are zero. They are not null values but they are absolute values. They indicate that we don’t have any questions at that time. Implying SARIMA for this particular subset will enable us to forecast with a better result. 

\begin{figure}[!h]
\centering
\includegraphics[width=0.5\textwidth]{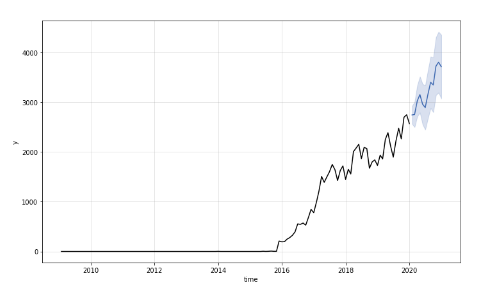}
\caption{\label{fig:deep-learning-framework-forecasting}Deep Learning Framework Forecasting}
\end{figure}

\begin{figure}[!h]
\centering
\includegraphics[width=0.9\textwidth]{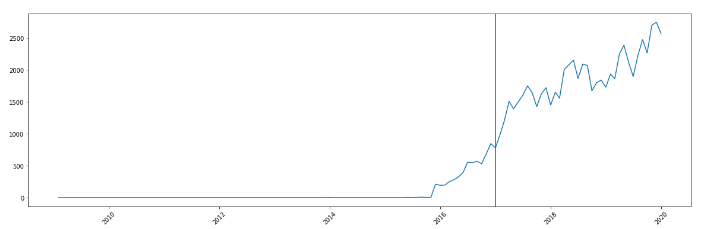}
\caption{\label{fig:point-of-change}Point of Change}
\end{figure}

\section{Error analysis}

The data is sliced and a part is used for future forecasting. The model predicts that there will be a total of 1721483 questions till 31/07/2021. It is estimated that 1,767,566 were available on that day. This is a result of one such technology. The table below represents the various technologies and their errors.

\begin{table}[h]

\begin{tabular}{|l|l|l|l|l|}
\hline
\textbf{Technology} & \textbf{Overall Error}  & \textbf{MAE}  & \textbf{MSE}   & \textbf{RMSE} \\ \hline
Python              & 1143.00                                       & 1112.56                       & 1737264.73                    & 1318.05                 \\ \hline
R                   & 500.00                                        & 669.90                        & 531123.78                     & 728.78                  \\ \hline
Pandas              & 262.07                                        & 166.40                        & 47888.59                      & 218.83                  \\ \hline
Numpy               & 118.95                                        & 115.87                        & 16764.73                      & 129.48                  \\ \hline
Keras               & 105.83                                        & 92.94                         & 10878.06                      & 104.30                  \\ \hline
Tensor              & 432.04                                        & 549.48                        & 335242.79                     & 579.00                  \\ \hline
Machine learning    & 179.64                                        & 177.22                        & 38562.49                      & 196.37                  \\ \hline
Deep learning       & 97.55                                         & 98.90                         & 12488.70                      & 111.75                  \\ \hline
OpenCV              & 72.96                                         & 74.67                         & 7081.20                       & 84.15                   \\ \hline
Apache-Spark        & 494.55                                        & 546.75                        & 318080.73                     & 563.99                  \\ \hline
\end{tabular}
\end{table}
\section{Limitations of the study}

The model that we designed so far, worked well on the small sample of data. The predictions of unseen data are quite a bit in correlation with the tuned model’s prediction. But, the limitations lie with the data set. Though the model’s forecasting is quite fit with reality, we have a very small volume of data. This will restrict us from further predictions. As of now we can predict accurately up to 2 years but when we can collect more data from different sources and communities we can put forward a strong prediction. This will help us in future forecasting. 

\section{Conclusion}

The basic ideology behind this paper is that we are targeting the stakeholders, companies and educationists to focus on what technology to use. Our predictions will help them to identify the best fit technology for their company and they can work with it. If they are working on technologies which are going down in trend they can switch. This will be applicable to students too. They can study trending technologies and frameworks to sustain in the IT industry. The future work will be focusing on training the model with a large volume of data and to fine tune the model. In addition to that, we can work on private company’s data to predict its trend. This will help them to emerge in new trending technologies as well as it improves their performance and efficiency. 

\section{Application}
We've developed an application on this too!\hfill \break
Visit \href{http://stackindex.tactii.com/}{http://stackindex.tactii.com/} to see how it works.

\section{Acknowledgements}

We would like to sincerely thank Vaishnavi Venkat, Aishwaraya Ramaswami, Raghul Kesavan, Sakthi Namasivayam, Akula Sainadh, Naveen V and Varun Abhishek for their work in data collection from Stackoverflow. 

\section{References}

[1] Samarth Tambad, Rohit Nandwani and Suzanne K. McIntosh “Analyzing programming languages by community characteristics on Github and StackOverflow” \hfill \break
[2] David Hin “StackOverflow vs Kaggle: A Study of Developer Discussions About Data Science ” \hfill \break
[3 ] Samyak Prajapati, Aman Swaraj, Ronak Lalwani, Akhil Narwal, Karan Verma, Ghanshyam Singh, Ashok Kumar “Comparison of Traditional and Hybrid Time Series Models for Forecasting COVID-19 Cases” \hfill \break
[4] June Young Lee, Sejung Ahn, 'Deep learning-based prediction of future growth potential of technologies' \hfill \break
[5] Jieun Kim and Christopher L. Magee 'Dynamic patterns of knowledge flows across technological domains: empirical results and link prediction'\hfill \break
[6]  Y. Tian, W. Ng, J. Cao and S. McIntosh, 'Geek talents: who are the top experts on github and stack overflow?,' Computers, Materials \& Continua \hfill \break
[7] Byungun Yoon, Christopher L.Magee 'Exploring technology opportunities by visualizing patent information based on generative topographic mapping and link prediction' \hfill \break
[8] Jaewoong Choia, Byeongki Jeong, Janghyeok Yoon, Byoung-YoulCoh, Jae-Min Lee 'A novel approach to evaluating the business potential of intellectual properties: A machine learning-based predictive analysis of patent lifetime' \hfill \break
[9]  Rojat, Thomas \& Puget, Raphaël \& Filliat, David \& Del Ser, Javier \& Gelin, Rodolphe \& Diaz Rodriguez, Natalia. (2021). Explainable Artificial Intelligence (XAI) on TimeSeries Data: A Survey. \hfill \break 
[10]  Sindhu Tipirneni, Chandan K. Reddy, “Self-supervised Transformer for Multivariate Clinical Time-Series with Missing Values” \hfill \break
[11] TIME SERIES ANALYSIS James D. Hamilton Princeton University Press, 1994 BE Hansen - Econometric Theory, 1995 \hfill \break
[12] Introduction to time series and forecasting PJ Brockwell, PJ Brockwell, RA Davis, RA Davis - 2016 \hfill \break
[13]  Firat, Ayse \& Woon, Wei \& Madnick, Stuart. (2012). “Technological Forecasting - A Review.” \hfill \break 
[14] J Lin, E Keogh, S Lonardi, B Chiu - Proceedings of the 8th ACM SIGMOD workshop on …, 2003 \hfill \break
[15] Won SangLee, Eun JinHan, So YoungSohn 'Predicting the pattern of technology convergence using big-data technology on large-scale triadic patents' \hfill \break
[16] Athiyarath, S., Paul, M. \& Krishnaswamy, S. A Comparative Study and Analysis of Time Series Forecasting Techniques. SN COMPUT. SCI. 1, 175 (2020). 25 years of time series forecasting \hfill \break
[17] JG De Gooijer, RJ Hyndman - International journal of forecasting, 2006  A symbolic representation of time series, with implications for streaming algorithms \hfill \break
[18] Agrawal RK, Adhikari R. An introductory study on time series modeling and forecasting. arXiv Preprint arXiv:1302.6613, 1302.6613 2013: 1-68. \hfill \break
[19] Anuraag Singh, Giorgio Triulzi, Christopher L. Magee 'Technological improvement rate predictions for all technologies' \hfill \break
[20] C. Chatfield, “Model uncertainty and forecast accuracy”, J. Forecasting 15 (1996) \hfill \break

\end{document}